\documentclass[letterpaper]{article} 
\usepackage{aaai25}  
\usepackage{times}  
\usepackage{helvet}  
\usepackage{courier}  
\usepackage[hyphens]{url}  
\usepackage{graphicx} 
\usepackage{natbib}  
\usepackage{caption} 
\usepackage{algorithm}

\usepackage{graphicx}
\usepackage{algorithmicx,algpseudocode}
\usepackage{caption}
\usepackage{xcolor}
\usepackage{amssymb}
\usepackage{mathtools}
\usepackage{mathabx}
\usepackage{comment}
\usepackage{graphicx}
\usepackage{booktabs}
\usepackage{enumitem}
\usepackage[skip=2.5pt]{caption}
\usepackage{multicol}
\usepackage{color, colortbl}

\usepackage[accsupp]{axessibility} 

\definecolor{Gray}{gray}{0.9}

\newcommand\given[1][]{\:#1\vert\:}

\DeclareMathOperator{\data}{data}

\newcommand\norm[1]{\left\lVert#1\right\rVert}
\newcommand{\overbar}[1]{\mkern 1.5mu\overline{\mkern-1.5mu#1\mkern-1.5mu}\mkern 1.5mu}

\newcommand{\argmin}{\mathop{\mathrm{argmin}}}

\newcommand{\Wc}{{\mathcal W}}

\newcommand{\xb}{{\boldsymbol x}}

\newcommand{\vb}{{\boldsymbol v}}

\newcommand{\epsilonb}{{\boldsymbol \epsilon}}

\newcommand{\Eb}{{\mathbb E}}

\newcommand{\local}{\text local}

\newcommand{\alphabar}{{\bar \alpha}}

\newcommand{\Ib}{{\boldsymbol I}}

\newcommand{\x}{{\boldsymbol x}}

\newcommand{\Nc}{{\mathcal N}}

\newcommand{\Pc}{{\mathcal P}}

\usepackage{newfloat}
\usepackage{listings}
\DeclareCaptionStyle{ruled}{labelfont=normalfont,labelsep=colon,strut=off} 
\floatstyle{ruled}
\newfloat{listing}{tb}{lst}{}
\floatname{listing}{Listing}
\title{Spectral Motion Alignment for Video Motion Transfer using Diffusion Models}
\author{
    Geon Yeong Park$^{1*}$, Hyeonho Jeong$^{2*}$, Sang Wan Lee$^{3\dag}$, Jong Chul Ye$^{1,2\dag}$ \\
}
\affiliations{
    $^{1}$Bio and Brain Engineering, $^{2}$Kim Jaechul Graduate School of AI, $^{3}$Brain and Cognitive Sciences \\
Korea Advanced Institute of Science and Technology (KAIST), Daejeon, Korea \\
$*, \dag$: co-first and co-corresponding authors. \\
{\tt\small \{pky3436, hyeonho.jeong, sangwan, jong.ye\}@kaist.ac.kr} 
}

\usepackage{bibentry}

\begin{document}

\twocolumn[{%
\renewcommand\twocolumn[1][]{#1}%
\maketitle
\vspace{-0.7cm}
\begin{center}
    \centering
    \captionsetup{type=figure}
    \includegraphics[width=\textwidth]
    {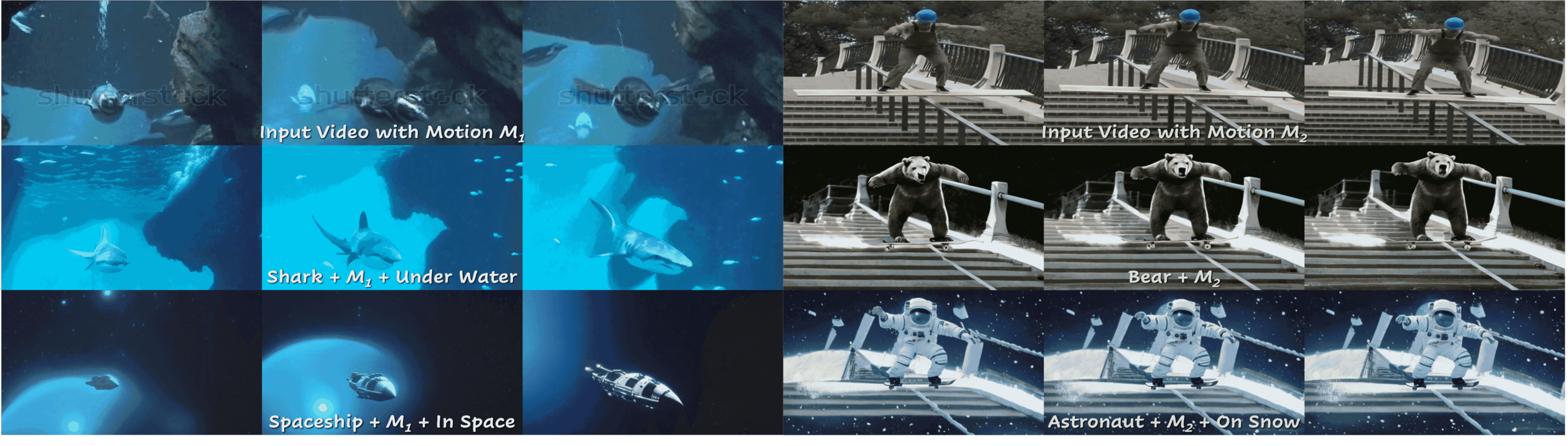}
    \captionof{figure}{
    One-shot Video Motion Transfer via \textbf{S}pectral \textbf{M}otion \textbf{A}lignment
    using Cascaded Video Diffusion Models. 
    \textbf{SMA} facilitates the capture of long-range (\textit{left}) and complex (\textit{right}) motion patterns within videos. 
    Visit \textit{https://geonyeong-park.github.io/spectral-motion-alignment/} for a comprehensive view of the videos.
    }
    \label{figure: teaser}
\end{center}
}]

\begin{abstract}
{Diffusion models have significantly facilitated the customization of input video with target appearance while maintaining its motion patterns. To distill the motion information from video frames, existing works often estimate motion representations as frame difference or correlation in pixel-/feature-space. Despite its simplicity, these methods have unexplored limitations, including lack of understanding of global motion context, and the introduction of motion-independent spatial distortions. To address this, we present {\em Spectral Motion Alignment (SMA)}, a novel framework that refines and aligns motion representations in the spectral domain. Specifically, SMA learns spectral motion representations, facilitating the learning of whole-frame global motion dynamics, and effectively mitigating motion-independent artifacts. Extensive experiments demonstrate SMA's efficacy in improving motion transfer while maintaining computational efficiency and compatibility across various video customization frameworks.}
\end{abstract}

%

\section{{Introduction}}
{Given the multifaceted nature of the video, encompassing motion dynamics, appearance, etc., several studies aim to disentangle and control these signals according to user intent. 
Recently, diffusion models  \cite{sohl2015deep, ho2020denoising} has played a pivotal role in video customization, owing to their superior sampling ability.}

{In the context of \textit{motion} customization using diffusion models, our goal is to transfer the motion patterns from an input video to the customized output video. This necessitates the accurate estimation and extraction of motion information from the input video. While fundamental techniques such as optical flow are effective for motion estimation, integrating these into diffusion models for customization is nontrivial.}

{To address these challenges, recent researches suggest that motion patterns are inherently encoded in the underlying dependencies between frames or epsilon noises. For example, \cite{zhao2023motiondirector} have observed that videos with similar motion tend to exhibit similar connectivity between latent frames. Additionally, \cite{jeong2023vmc} utilizes residual vectors between consecutive frames as "motion vectors," in line with optical flow principles, assuming frame residuals represent motion dynamics. 
Specifically, they finetune pretrained VDM to align the ground-truth pixel-space residuals with their predicted denoised estimates. Thus, these works leverage the pixel-space differences between input frames as a proxy of motion reference.}

{While these motion representations can be obtained from off-the-shelf video diffusion models efficiently, current simple approximations have several adverse impacts. First, they may fail to capture the global context of motion. Since frame residuals may capture local motion patterns but are blind to whole-frame motion dynamics, for better motion dynamics modeling, we have to understand the whole-frame global context information during motion distillation. Furthermore, while the pixel- or feature-space residuals contain motion information, they may also contain inevitable disruptive variations that are unrelated to motion. These variations may include abrupt changes in the background, lightning, or other frame inconsistencies, leading to less reliable representation.}

{To address these challenges, we introduce Spectral Motion Alignment (\textbf{SMA}), a novel framework for refining and aligning motion representations in the \textit{spectral} domain, based on intuition that the motion may be well represented by its inherent frequency components. 
This framework includes two primary components: First, to capture the global motion context, we propose a spectral alignment loss between predicted and ground-truth motion vectors within the wavelet domain. This facilitates the learning of multi-scale motion dynamics by leveraging rich wavelet-domain representations of video considering the global frame transitions. Moreover, to mitigate the spatial artifacts and inconsistency in motion vectors, we propose 2D FFT-based motion vector refinement that aligns the amplitude and phase spectrum of ground truth and predicted motion vectors with prioritizing low-frequency components. This is because the high-frequency components in motion representations may be associated with frame-wise motion-independent artifacts (Figure \ref{fig: local ablation}). In summary, we encourage accurate motion transfer via harmonized global and local levels of spectral domain alignment. Our contributions are summarized as follows:}

\begin{figure*}[t]
    \centering
    \includegraphics[width=\textwidth]{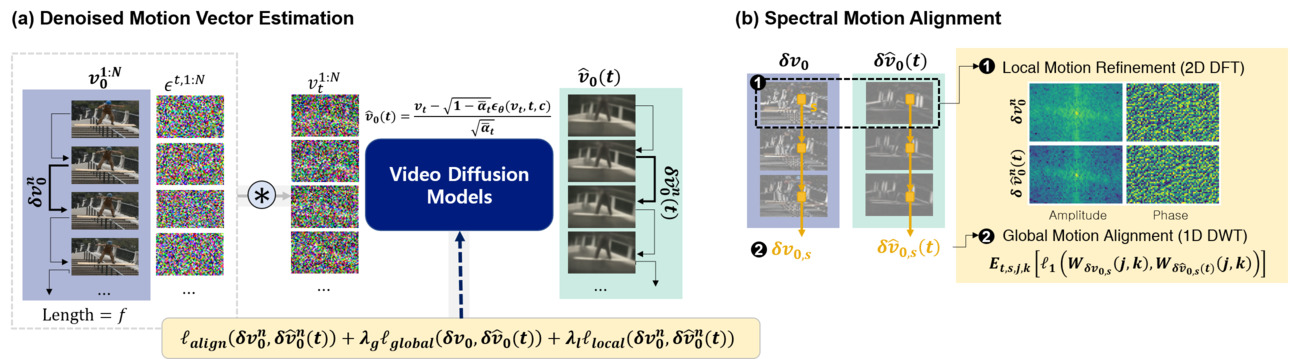}
    \caption{\textbf{Overview.} The proposed Spectral Motion Alignment (\textbf{SMA}) framework distills the motion information in frequency-domain. Considering the (latent) frame residuals as motion vectors, we first derive the denoised motion vector estimates. Then, the motion vector $\delta \vb_0^n$ and its estimate $\delta \hat{\vb}_0^n$ are aligned in both pixel-domain and frequency-domain. Our regularization includes (1) global motion alignment based on 1D wavelet-transform, and (2) local motion refinement based on 2D Fourier transform.}
    \vspace{-0.2cm}
    \label{fig:enter-label}
\end{figure*}

\begin{itemize}[noitemsep]
    \item {We introduce the Spectral Motion Alignment (SMA), a frequency-domain motion alignment framework that learns the underlying motion dynamics of input video via frequency-based regularization. SMA is orthogonal and compatible to most of existing motion customization models as they often only rely on either pixel or feature space representations.}
    \item SMA imposes negligible memory and computational burdens, as most off-the-shelf VDMs can readily compute motion vectors estimates. For instance, VMC \cite{jeong2023vmc} with SMA demonstrates lightweight (15GB vRAM) and rapid ($<$ 5 min) training.
    \item We validate the efficacy of SMA across diverse motion patterns, subjects, and various video motion transfer frameworks including Video Diffusion-based \cite{zhao2023motiondirector}, Cascaded Video Diffusion-based \cite{jeong2023vmc}, T2I Diffusion-based \cite{wu2023tune}, and ControlNet-based models \cite{chen2023control}.
\end{itemize}
\vspace{-3pt}

\section{Preliminaries}
\label{sec: prelim}

\subsection{Diffusion Models.} 
Diffusion models \cite{sohl2015deep, ho2020denoising} generate samples from the Gaussian noise through reverse denoising processes. We denote a clean sample $\xb_0 \sim p_{\data}(\xb)$, a noisy latent $\xb_{t} \in \mathbb{R}^d$ at time $t$, $\beta_t$ as an increasing sequence of noise schedule, $\alpha_t \coloneqq 1 - \beta_t$, and $\alphabar_t \coloneqq \Pi_{i=1}^t \alpha_i$. Then the goal of diffusion model training is to optimize a denoiser $\epsilonb_{\theta^*}$: 
\begin{equation}
    \label{eq: epsilon matching}
   \theta^*:=\argmin_{\theta} \Eb_{\xb_t, \xb_0, \epsilonb} \big[ \norm{\epsilonb_{\theta} (\xb_t, t) - \epsilonb} \big] .
\end{equation}

\noindent The reverse sampling from $q(\xb_{t-1}|\xb_t, \epsilonb_{\theta^*}(\xb_t, t))$ is then achieved by
\begin{align}
    \label{eq: reverse sampling}
    \x_{t-1}=\frac{1}{\sqrt{\alpha_t}}\Big{(}\x_t-\frac{1-\alpha_t}{\sqrt{1-\bar{\alpha}_t}}\boldsymbol{\epsilon}_{\theta^*}(\x_t, t)\Big{)}+\tilde{\beta}_t \epsilonb,
\end{align}
where $\epsilonb \sim\Nc(0,\Ib)$ and $\tilde{\beta}_t \coloneqq \frac{1 - \alphabar_{t-1}}{1 - \alphabar_t} \beta_t$. 
To accelerate sampling, DDIM \cite{song2020denoising} further proposes another sampling method as follows:
\begin{equation}
    \label{eq: ddim sampling}
    \xb_{t-1} = \sqrt{\alphabar_{t-1}} \hat{\xb}_0(t) + \sqrt{ 1 - \alphabar_{t-1} - \eta^2 \tilde{\beta_t}^2} \epsilonb_{\theta^*} (\xb_t, t) + \eta \tilde{\beta}_t \epsilonb, \nonumber
\end{equation}
where $\eta \in [0,1]$ controls stochasticity, and $\hat{\xb}_0(t)$ is the denoised estimate which can be equivalently derived using Tweedie's formula \cite{efron2011tweedie}:
\begin{equation}
    \label{eq: Tweedie}
    \hat{\xb}_0(t) \coloneqq \frac{1}{\sqrt{\alphabar_t}} (\xb_t - \sqrt{1-\alphabar_t} \epsilonb_{\theta^*} (\xb_t, t)).
\end{equation}
For a text-guided generation, diffusion models are often trained with the textual embedding $c$. Throughout this paper, we will often omit $c$ from $\epsilonb_{\theta} (\xb_t, t, c)$ if it does not lead to notational ambiguity.

\subsubsection{Video Diffusion Models.} 
Video diffusion models \cite{ho2022video, ho2022imagen, zhang2023show} further attempt to model the video data distribution. Specifically, Let $(\vb^n)_{n \in \{1, \dots, N\}}$ represents the $N$-frame input video sequence. Then, for a given $n$-th frame $\vb^n \in \mathbb{R}^d$, let $\vb^{1:N} \in \mathbb{R}^{N \times d}$ represents a whole video vector. Let $\boldsymbol{v}_t^n = \sqrt{\bar{\alpha}_t} \boldsymbol{v}^n + \sqrt{1 - \bar{\alpha}_t} \boldsymbol{\epsilon}_t^{n}$ represents the $n$-th noisy frame latent sampled from $p_t(\boldsymbol{v}_t^n | \boldsymbol{v}^n)$, where $\boldsymbol{\epsilon}_t^n \sim \mathcal{N}(0, I)$. We similarly define $(\vb_t^n)_{n \in 1, \dots, N}$, $\vb_t^{1:N}$, and $\epsilonb^{1:N}$. The goal of video diffusion model training is then to obtain a residual denoiser $\epsilonb_{\theta} $ with textual condition $c$ and video input that satisfies:
\begin{equation}
    \label{eq: video epsilon matching}
    \min_{\theta} \Eb_{\vb_t^{1:N}, \vb^{1:N}, \epsilonb^{1:N}, c} \big[ \norm{\epsilonb_{\theta} (\vb_t^{1:N}, t, c) - \epsilonb^{1:N}} \big],
\end{equation}
where $\epsilonb_\theta (\vb_t^{1:N}, t, c), \epsilonb^{1:N} \in \mathbb{R}^{N \times d}$. In this work, we denote the predicted noise of $n$-th frame as $\epsilonb_\theta^n (\vb_t^{1:N}, t, c) \in \mathbb{R}^d$.


\subsection{Fourier and Wavelet Analysis} 
Spectral analysis techniques transform time-domain or pixel-domain signals (such as video frames) into the frequency domain, revealing the frequency components and their intensities. 

\noindent \textbf{Fourier Transform.} Let $\vb^n \in \mathbb{R}^{H \times W}$ represents the $n$-th 2D video frame. Then, its frequency spectrum at coordinate $(a, b)$ is given as follows:

\begin{equation}
\label{eq: 2d dft}
    \mathcal{F}_{\vb^n} (a, b) = \sum_{x=0}^{H-1} \sum_{y=0}^{W-1} \vb^n(x, y) e^{-i 2\pi (\frac{ax}{H} + \frac{by}{W})},
\end{equation}
where $\vb^n(x, y)$ means the pixel value at coordinate $(x, y)$. The output frequency spectrum is represented as $\mathcal{F}_{\vb^n} (a, b) = R(a, b) + I(a, b)i$, where $R(a, b), I(a, b) \in \mathbb{R}$ represents real and imaginary part, respectively. Then, the amplitude and phase is derived as follows:
\begin{equation}
\begin{split}
    | \mathcal{F}_{\vb^n} (a, b) | = \sqrt{R(a, b)^2 + I(a, b)^2}, \\
    \angle \mathcal{F}_{\vb^n} (a, b) = \arctan \Big( \frac{I(a,b)}{R(a,b)} \Big).
\end{split}
\label{eq: amplitude and phase}
\end{equation}

\noindent \textbf{Wavelet Transform.} Wavelet frames, renowned for capturing multi-resolution scale features, are among the most prevalently utilized frame representations in signal processing. Let $\psi(t)$ represent a mother wavelet that can be shifted and scaled. For a function $\vb(t) \in L^2(\mathbb{R})$, the wavelet transform can be expressed as:

\begin{equation}
\label{eq: continuous WT}
    \mathcal{CW}_{\vb}(a, b) = \frac{1}{\sqrt{\alpha}} \int \vb(t) \psi^* \left(\frac{t-b}{a} \right)dt = \langle \vb(t), \psi_{a,b}(t) \rangle,
\end{equation}
which serves as an expansion coefficient. 
In the case of discrete wavelet transform (DWT), it uses a finite set of wavelet and scaling functions derived from a chosen wavelet family. Specifically, the mother wavelet is shifted and scaled by powers of two as follows: 
\begin{equation}
    \label{eq: scaled wavelet}
    \psi_{j, k}(t) = \frac{1}{\sqrt{2^j}} \psi(2^{-j}t - k).
\end{equation}
Then, the DWT of a signal $\vb[n]$ is given by:
\begin{equation}
    \label{eq: discrete WT}
    \mathcal{W}_{\vb}(j, k) =  \langle \vb(t), \psi_{j, k}(t) \rangle.
\end{equation}
The original signal can be recovered from inverse DWT. In practice, this discrete wavelet transform can be implemented by convolution using an appropriate choice of filter bank.
\vspace{-0.2cm}

\section{Spectral Motion Alignment}
{Our main goal is to develop a novel \textit{spectral domain} motion alignment framework that capture underlying complex motion patterns across a spectrum of frequency levels that mainly constitute motion. This is valuable in video understanding and customization as it helps in identifying repetitive motion patterns and underlying structures that may not be visible in the time or pixel domain. It is in orthogonal (and compatible) with conventional methods based on pixel- or feature-domain motion representations.} 

\vspace{-0.2cm}
\subsection{Denoised Motion Vector Estimation}
{To distill the motion information, we first estimate the initial motion representations \cite{jeong2023vmc, zhao2023motiondirector} in pixel space. For this, we follow VMC \cite{jeong2023vmc} as an representative example. The intuition is that residual vectors between consecutive frames may include information about the motion trajectories. Define the $n$-th frame residual vector, namely motion vector at time $t \geq 0$ as} 

\begin{equation}
    \label{eq: frame residual vector}
    \delta \vb_t^n \coloneqq \vb_t^{n+1} - \vb_t^n,
\end{equation}
where the epsilon residual vector $\delta \epsilonb_t^n$ is similarly defined. This $\delta \vb_t^n$ can be acquired through the following diffusion kernel:
\begin{equation}
    \label{eq: residual frame kernel}
    p(\delta \vb_t^n \given \delta \vb_0^n) = \Nc (\delta \vb_t^n \given \sqrt{\alphabar_t} \delta \vb_0^{n}, 2(1 - \alphabar_t)I).
\end{equation}

\noindent Given that, the ground-truth motion vector in pixel space $\delta \vb_0^n$ can be derived as follows:
\begin{equation}
    \label{eq: ground-truth motion vector}
    \delta \vb_0^n = \frac{1}{\sqrt{\alphabar_t}} \Big( \delta \vb_t^n - \sqrt{1-\alphabar_t} \delta \epsilonb_t^n \Big).
\end{equation}

\noindent Similarly, one can obtain the denoised estimate version of these motion representations $\delta \hat{\vb}_0^n$ by using Tweedie's formula as follows: 
\begin{equation}
    \label{eq: video Tweedie}
    \hat{\vb}_0^{1:N}(t) \coloneqq \frac{1}{\sqrt{\alphabar_t}} \big( \vb_t^{1:N} - \sqrt{1-\alphabar_t} \epsilonb_{\theta} (\vb_t^{1:N}, t) \big),
\end{equation}
where $\hat{\vb}_0^{1:N}(t)$ is an empirical Bayes optimal posterior expectation $\mathbb{E} [\vb_0^{1:N} \given \vb_t^{1:N}]$.

In the context of motion transfer, we aim to align the ground-truth and estimated motion vectors by fine-tuning the pre-trained VDM \cite{jeong2023vmc, zhao2023motiondirector}:
\begin{equation}
    \label{eq: v0 align}
    \min_{\theta} \mathbb{E}_{t, n, \epsilonb^{t, n}, \epsilonb^{t, n+1}} \Big[\ell_{\text{align}} \big(\delta \vb_0^n, \delta \hat{\vb}_0^n(t) \big) \Big].
\end{equation}
While these advancements in motion distillation mark significant progress, Figure \ref{figure: teaser}, \ref{fig:comparison-vmc} and \ref{fig: local ablation} indicate that existing methods still has potential for further refinement.

\vspace{-0.1cm}
\subsection{Spectral Global Motion Alignment}
One of the primary limitations in \eqref{eq: v0 align} is that it may not fully encapsulate the global motion dynamics. Specifically, it locally focuses on pairwise frame comparisons which may lead to overlooking the comprehensive motion dynamics of given object overall frames. 

To mitigate these problems, we explore the use of wavelet transforms in motion distillation. In this paper, we use Haar wavelet, whose low and high pass filters are given as follows:
\begin{equation}
    \label{eq: haar}
 L[n] = \frac{1}{\sqrt{2}} [1 \ 1], H[n] = \frac{1}{\sqrt{2}} [-1 \ 1],
\end{equation}
\noindent which is implemented using the multi-scale Haar filter bank. Then, given the sequence of motion vectors $\delta \vb_0=(\delta \vb_0^n)_{n\in\{1, \dots, N-1\}}$ and its denoised estimates $\delta \hat{\vb}_0(t)=\big( \delta \hat{\vb}_0^n(t) \big)_{n\in \{1, \dots, N-1\}}$, we consider ($N-1$)-length time-dependent 1D arrays from arbitrary spatial pixel dimension $s \in \{1, \dots d\}$. The corresponding 1D array of motion vector is denoted by $\delta \vb_{0, s}$ and $\delta \hat{\vb}_{0, s}(t) \in \mathbb{R}^{N - 1}$ (Figure \ref{fig:enter-label}).

Then, the frequency-matching loss between $\delta \vb_0$ and $\delta \hat{\vb}_0(t)$ is defined with DWT in \eqref{eq: discrete WT} as follows:
\begin{equation}
\begin{split}
\label{eq: global matching loss}
    \ell_{\text{global}}&(\delta \vb_0, \delta \hat{\vb}_0(t)) = \\
    &\mathbb{E}_{t, s, j, k} \Big[ \| \Wc_{\delta \vb_{0, s}}(j, k) - \Wc_{\delta \hat{\vb}_{0, s}(t)}(j, k) \|_1 \Big].
\end{split}
\end{equation}
Considering that the wavelet transform allows multi-resolution analysis of motion vectors, it enables us to handle motions at various scales and frequencies effectively. This could be particularly beneficial for complex scenes with varying motion speeds and types, ensuring that subtle motions are captured and transferred more accurately. 

\vspace{-0.1cm}
\subsection{Spectral Local Motion Refinement}
Another problem in \eqref{eq: v0 align} is that the estimated motion representations may encapsulate high-frequency local distortions, background noise, and other non-motion-related artifacts. By aligning the denoised estimates with these artifacts, the fine-tuned VDM may erroneously reproduce similar high-frequency artifacts as in Figure \ref{fig: local ablation}.

Accordingly, we 
focus on prioritizing low-to-moderate spatial frequency components particularly. 
Specifically, following the amplitude and phase spectrum definition in \eqref{eq: amplitude and phase}, we define amplitude and phase matching loss, $\ell_{\local}^{A}(\delta \vb_0^n, \delta \hat{\vb}_0^n(t))$ and $\ell_{\local}^{P}(\delta \vb_0^n, \delta \hat{\vb}_0^n(t))$, as follows:
\begin{equation}
\label{eq: amplitude matching loss}
\begin{split}
    &(A) \ \mathbb{E}_{t, n, a, b} \Big[ \omega(a, b) \| |\mathcal{F}_{\delta \vb_0^n} (a, b)| - |\mathcal{F}_{\delta \hat{\vb}_0^n(t)} (a, b)| \|_1  \Big], \\
    &(P) \ \mathbb{E}_{t, n, a, b} \Big[ \omega(a, b) \| \angle \mathcal{F}_{\delta \vb_0^n} (a, b) - \angle \mathcal{F}_{\delta \hat{\vb}_0^n(t)} (a, b) \|_1  \Big],
\end{split}
\end{equation}
where the frequency domain weighting $\omega(a,b)$ is defined as
\begin{equation}
\label{eq: low-freq weight}
\begin{split}
    \omega(a, b) = \Big[ \big(\frac{H}{2}\big)^2 + \big(\frac{W}{2}\big)^2 \Big]^\delta - \Big[ \big(a - \frac{H}{2}\big)^2 + \big(b - \frac{W}{2}\big)^2 \Big]^\delta+ 1 \nonumber
\end{split}
\end{equation}
for $0 < a < H, 0 < b < W$, and otherwise, set to zero.
This introduces a weighting \cite{yang2022delving} that prioritizes low-frequency components for $\delta > 0$.

\vspace{-0.2cm}
\begin{figure*}[!t]
    \centering
    \includegraphics[width=\textwidth]{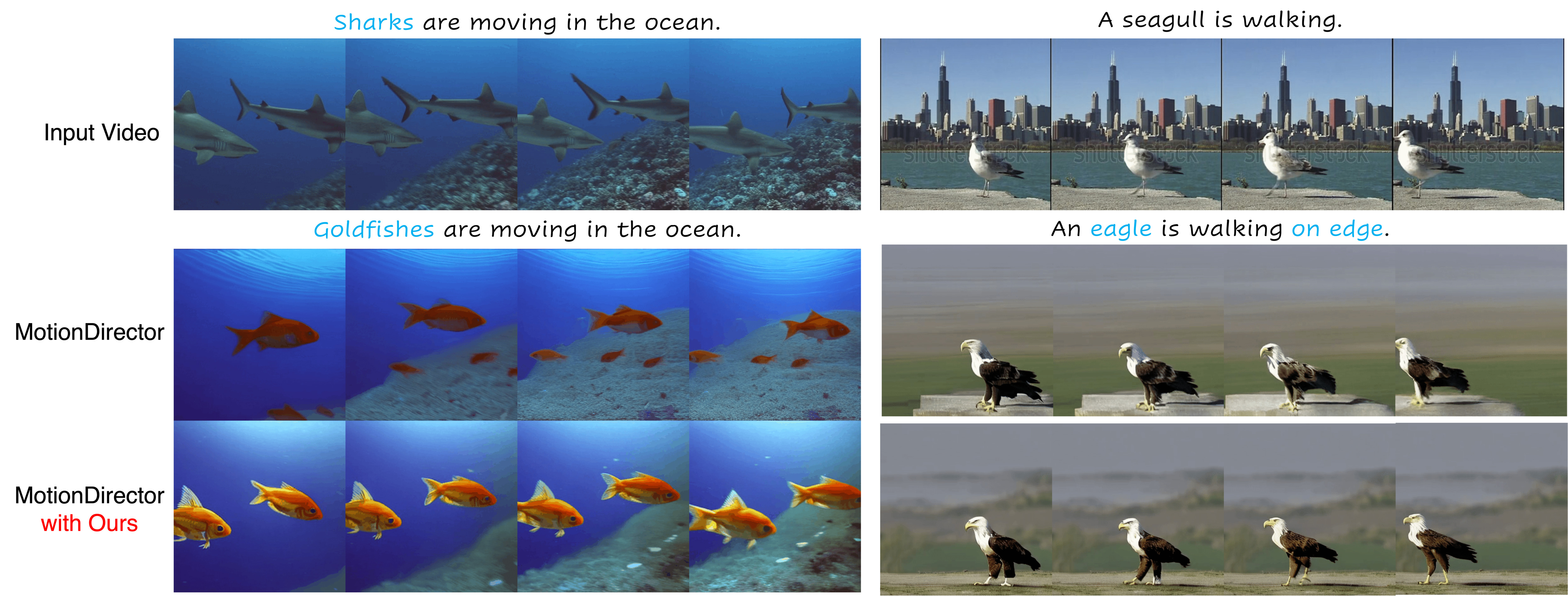}
    \caption{Comparison within MotionDirector framework.}
    \label{fig:comparison-md}
\end{figure*}

\subsection{Inference Pipeline}
To sum up, the overall spectral motion alignment framework is given as follows:

\begin{equation}
\begin{split}
    \min_\theta &\mathbb E_{t, n, \epsilonb_t^n, \epsilonb_t^{n+1}} \Big[ \ell_{\text{align}} \big(\delta \vb_0^n, \delta \hat{\vb}_0^n(t)\big) + \\ 
    &\lambda_{g} \ell_{\text{global}} \big( \delta \vb_0, \delta \hat{\vb}_0(t) \big) + \lambda_{l} \ell_{\local} \big( \delta \vb_0^n, \delta \hat{\vb}_0^n(t) \big) \Big],
\end{split}
\end{equation}
where $\ell_{\local}(\delta \vb_0^n, \delta \hat{\vb}_0^n(t)) = \ell_{\local}^A(\delta \vb_0^n, \delta \hat{\vb}_0^n(t)) + \ell_{\local}^P(\delta \vb_0^n, \delta \hat{\vb}_0^n(t))$.
Upon optimization, the inference is performed using new text prompts to transform appearances, e.g. $\text{"a seagull is walking"} \rightarrow \text{"a chicken is walking"}$. 

This Spectral Motion Alignment (SMA) is universally adaptable across various motion distillation frameworks. While diverse diffusion-based motion distillation frameworks adopt their pixel-domain motion learning objectives, the proposed frequency-domain alignment seamlessly integrates with these arbitrary objectives. Moreover, different motion distillation frameworks target specific parameters $\theta$ for fine-tuning, varying from temporal attention layers \cite{jeong2023vmc} to dual-path LoRAs \cite{zhao2023motiondirector}. We empirically demonstrate the global compatibility of the proposed spectral motion alignment with diverse neural architectures and parameterizations. Pseudo-code is provided in the appendix. 

\subsection{{Extending SMA to Feature Space}}
\label{sec: dmt}
\begin{figure*}[!t]
    \centering
    \includegraphics[width=\textwidth]{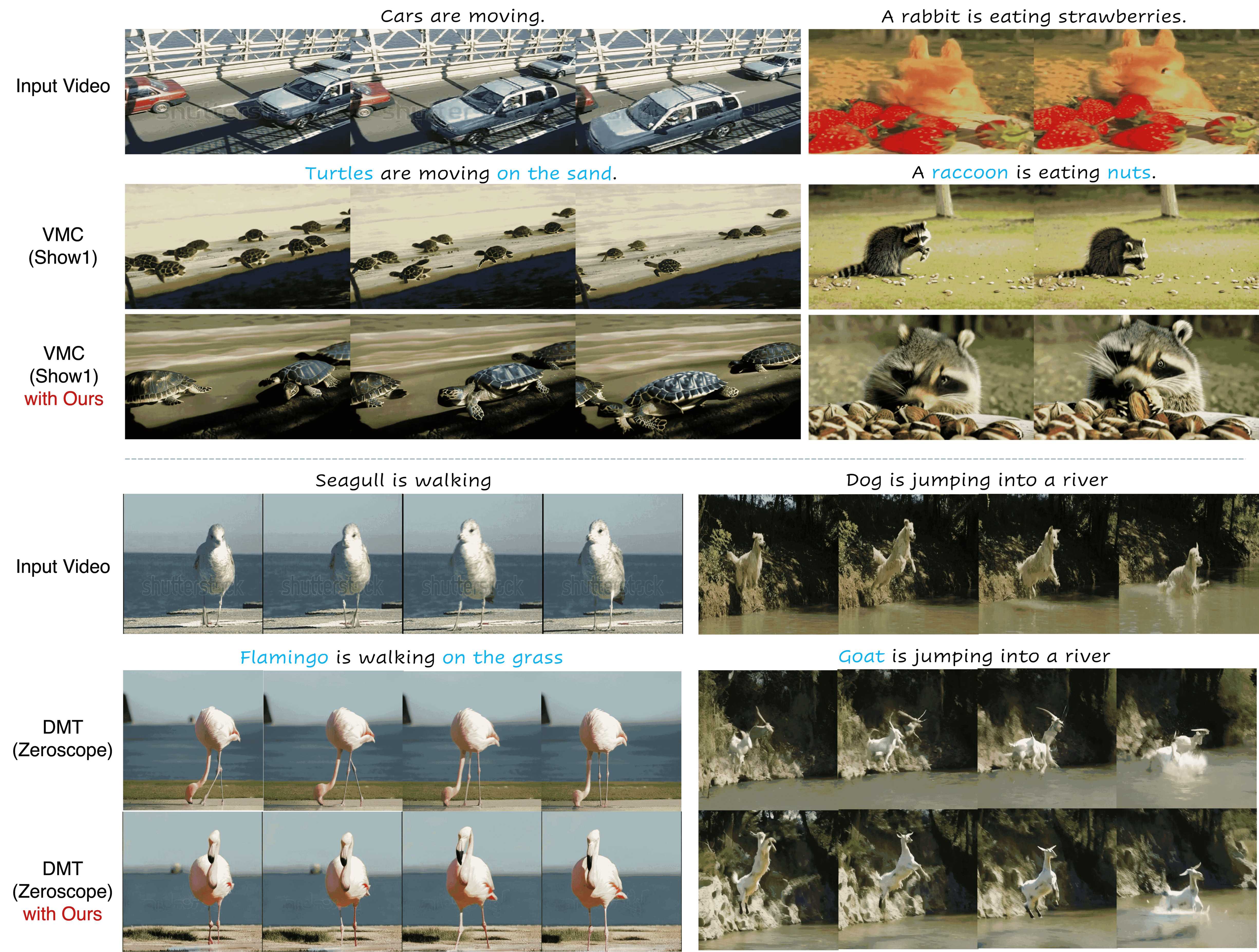}
    \caption{{Comparison within VMC framework using Show-1 video model (\textit{top}) and DMT framework using Zeroscope video model (\textit{bottom}). Each demonstrate the compatibility of SMA in pixel-space and feature-space, respectively.}}
    \vspace{-0.2cm}
    \label{fig:comparison-vmc}
\end{figure*}

{Beyond pixel-space motion representations, SMA can be further extended to semantic diffusion features (DIFT, \cite{tang2023emergent}). Specifically, Diffusion-Motion-Transfer (DMT, \cite{yatim2023space}) constructs motion vectors based on pairwise differences in space-time diffusion features, which are then utilized for latent optimization-based video motion transfer.
Given input and target video latents, $\vb_t$ and $\tilde{\vb}_t$, the model extracts space-time features $f(\vb_t)$ and $f(\tilde{\vb}_t)$. Then, the feature residuals are defined as $\delta f(\vb_t)^n$ and $\delta f(\tilde{\vb}_t)^n$, the $n$-th consecutive difference between hidden feature frames. This leads to the spectral alignment objective in feature space as follows:}
\begin{equation}
\begin{split}
\label{eq: dmt}
    \mathbb{E} \Big[&\ell_{\text{DMT}} \big( f(\vb_t),  f(\tilde{\vb}_t) \big) + \\
    &\lambda_{g} \ell_{\text{global}} \big( \delta f(\vb_t), \delta f(\tilde{\vb}_t) \big) + \lambda_{l} \ell_{\local} \big( \delta f(\vb_t)^n, \delta f(\tilde{\vb}_t)^n \big) \Big],
\end{split}
\end{equation}
where $\ell_{\text{DMT}}$ refers to the original space-time feature loss in \cite{yatim2023space}. Note that DMT does not finetune the models, leveraging \eqref{eq: dmt} for a latent optimization in sampling process.
We demonstrate the effectiveness of spectral alignment in the diffusion feature space by comparing it against the original DMT framework in Fig.\ref{fig:comparison-vmc}-bottom.


\section{Experiments using T2V Diffusion Models}
\label{exp: t2v}

\subsection{Experimental Setting}
\label{exp: t2v-setting}
To assess the capability of Spectral Motion Alignment (SMA) to capture accurate motion within contemporary diffusion-based motion learning frameworks, we curated a dataset comprising 30 text-video pairs sourced from the publicly available DAVIS \cite{pont20172017} and WebVid-10M \cite{Bain21} collections.
This dataset is deliberately designed to cover a broad spectrum of motion types and subjects, with video lengths ranging between 8 and 16 frames.
For this study, we leverage two foundational text-to-video diffusion models: Zeroscope \cite{zeroscope}, a non-cascaded VDM, and Show-1 \cite{zhang2023show}, a cascaded VDM. More details are provided in appendix.
\vspace{-0.1cm}

\subsection{Baselines}
\label{exp: motiondirector}
\textbf{MotionDirector} \cite{zhao2023motiondirector} tailor the appearance and motion of a video by developing a unique dual-path (spatial, temporal) framework with Low-Rank Adaptation (LoRA, \cite{hu2021lora}). \textbf{VMC} \cite{jeong2023vmc} achieves state-of-the-art performance in motion customization through their novel epsilon residual matching objective, facilitating efficient motion distillation within a cascaded video diffusion. \textbf{DMT} \cite{yatim2023space} proposes a new space-time feature loss, guiding the sampling process towards preserving the motion patterns while complying with the target object. Please see Sec \ref{sec: dmt} for more details.


\subsection{Qualitative Comparison.}
\label{sec: qualitative-t2v}
Fig. \ref{fig:comparison-md} and \ref{fig:comparison-vmc} offer qualitative comparisons with and without SMA.
The top of Figure \ref{fig:comparison-vmc} shows videos from a cascaded diffusion pipeline, while the bottom displays those from a non-cascaded model. Without SMA, videos may capture appearance to some extent but fail to replicate motion patterns accurately. In contrast, SMA significantly improves motion transfer, distinguishing dynamic from static objects.
For instance, in the last example of Fig. \ref{fig:comparison-md}, SMA produces a video where only the eagle moves from right to left, whereas without SMA, the video inaccurately depicts the ground moving alongside the eagle.

\subsection{Quantitative Comparison.}
\label{sec: quantiative-t2v}
The results of our quantitative evaluation are presented in Table \ref{quantitative-t2v}.
To evaluate text-video alignment \cite{hessel2021clipscore}, we measure the average cosine similarity between the target text prompt and the frames generated. 
Regarding frame consistency, we extract CLIP image features for each frame in the output video and subsequently calculate the average cosine similarity among all frame pairs in the video.
For human evaluation, we conduct a user study with 42 participants to assess three key aspects, guided by the following questions:
(1) Editing Accuracy: \textit{Is the output video accurately edited, reflecting the target text?}
(2) Temporal Consistency: \textit{Is the transition between frames smooth and consistent?}
(3) Motion Accuracy: \textit{Is the motion of the input video accurately preserved in the output video?}
Tab. \ref{quantitative-t2v} demonstrates that SMA enhances the performance of MotionDirector and VMC across all measured metrics.

\begin{table}
\centering
\resizebox{\linewidth}{!}{
\begin{tabular}{@{\extracolsep{0pt}}cccccc@{}}
\toprule
& \multicolumn{2}{c}{Automatic Metrics} & \multicolumn{3}{c}{User Study} \\
\cmidrule(lr){2-3} \cmidrule(lr){4-6}
Method              & Text-Align & Temp-Con & Edit-Acc & Temp-Con & Motion-Acc \\
\cmidrule{1-6}
MotionDirector          & 0.7550 & 0.9780 & 3.19 & 3.07 & 2.67 \\
\rowcolor{Gray}         
MotionDirector w/ Ours  & \textbf{0.8081} & \textbf{0.9784} & \textbf{4.14} & \textbf{3.89} & \textbf{3.88} \\
\cmidrule{1-6}
VMC (Show-1)            & 0.8066 & 0.9742 & 3.25 & 3.22 & 2.72 \\
\rowcolor{Gray}         
VMC (Show-1) w/ Ours    & \textbf{0.8193} & \textbf{0.9776} & \textbf{3.85} & \textbf{3.84} & \textbf{3.92} \\
\cmidrule{1-6}
VMC (Zeroscope)         & 0.8223 & 0.9560 & 2.95 & 2.83 & 2.47 \\
\rowcolor{Gray}         
VMC (Zeroscope) w/ Ours & \textbf{0.8425} & \textbf{0.9578} & \textbf{4.07} & \textbf{3.84} & \textbf{4.07} \\
\bottomrule
\end{tabular}
}
\caption{
Quantitative evaluation of SMA within text-to-video based frameworks.
}
\vspace{-0.3cm}
\label{quantitative-t2v}
\end{table}

\begin{figure}[!t]
   \centering
   \includegraphics[width=\columnwidth]{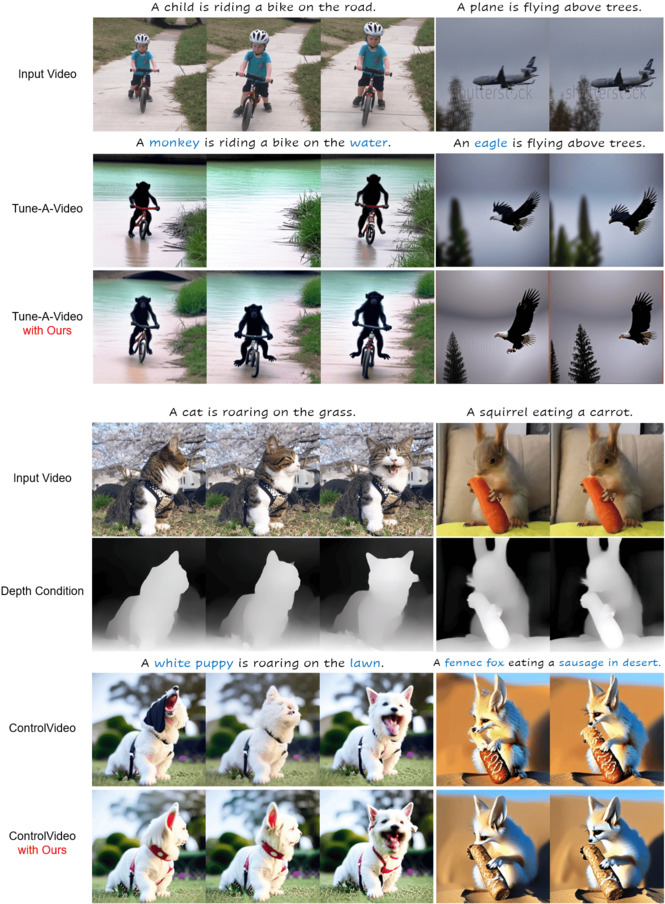}
   \caption{Comparison within Tune-A-Video (Top) and ControlVideo-Depth (Bottom) baseline.}
   \vspace{-0.5cm}
   \label{fig:comparison-tav-cv}
\end{figure}

\vspace{-0.2cm}
\begin{figure*}[htbp]
    \centering
    \includegraphics[width=0.92\textwidth]{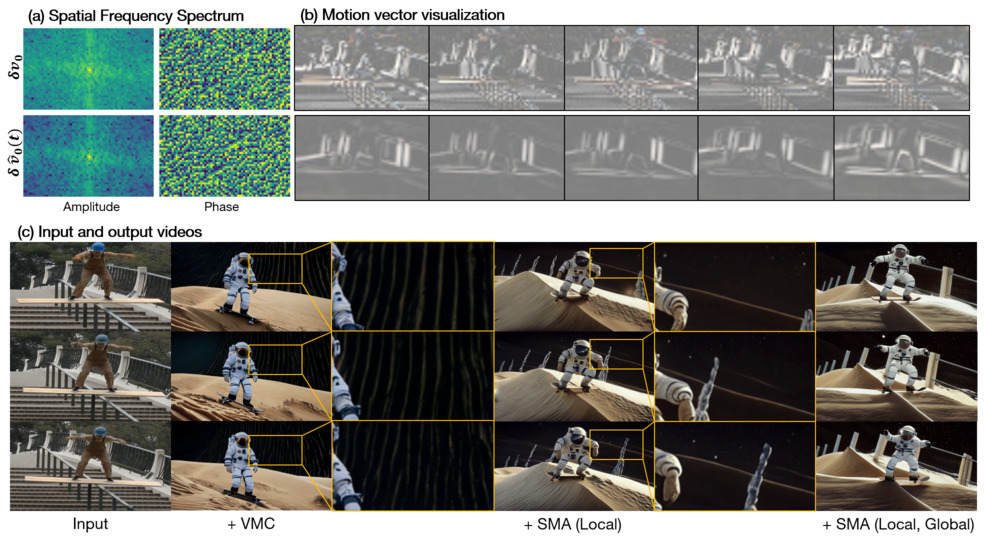}
    \caption{Visualization of (\textbf{a}) spatial frequency spectrum and (\textbf{b}) motion vectors estimated from the pre-trained Show-1 \cite{zhang2023show} without fine-tuning. (\textbf{c}) Ablation study on spectral motion alignment based on VMC \cite{jeong2023vmc}.}
    \label{fig: local ablation}
    \vspace{-0.4cm}
\end{figure*}

\section{T2I Video Diffusion Models}
\label{exp: t2i}

\subsection{Experimental Setting}
\label{exp: t2i-setting}
We further evaluate the efficacy of SMA with methods based on text-to-image diffusion model. Same text-video pairs are used as in Sec. \ref{exp: t2v-setting}.
The resolution for all produced videos is standardized to 512x512.
In this experiment, Stable Diffusion v1-5 \cite{rombach2022high} and ControlNet-Depth \cite{zhang2023adding} are utilized.

\subsection{Baselines}
\textbf{Tune-A-Video }\cite{wu2023tune} transforms a pretrained T2I diffusion model to psuedo T2V model by adding temporal attention layers and expanding spatial self-attention into spatio-temporal attention.
\textbf{ControlVideo} \cite{zhao2023controlvideo} is another one-shot-based video editing method stems from pretrained T2I model. 
ControlVideo extends ControlNet \cite{zhang2023adding} from image to video to incorporate structural cues obtained from the input video.

\subsection{Qualitative Comparison.}
Fig. \ref{fig:comparison-tav-cv}-(\textit{top}) demonstrates the efficacy of SMA with Tune-A-Video method, where SMA alleviates the flickering artifacts in foreground objects.
Fig. \ref{fig:comparison-tav-cv}-(\textit{bottom}) further illustrates the improvements with the ControlVideo framework.
While depth control encourages ControlVideo to maintain the structural integrity, Fig. \ref{fig:comparison-tav-cv} shows that it is not sufficient for motion accuracy, where SMA plays a crucial role in accurate capture of motion details.

\begin{table}
\centering
\resizebox{0.95\linewidth}{!}{
\begin{tabular}{@{\extracolsep{0pt}}cccccc@{}}
\toprule
& \multicolumn{2}{c}{Automatic Metrics} & \multicolumn{3}{c}{User Study} \\
\cmidrule(lr){2-3} \cmidrule(lr){4-6}
Method              & Text-Align & Temp-Con & Edit-Acc & Temp-Con & Motion-Acc \\
\cmidrule{1-6}
Tune-A-Video           & 0.8289 & 0.8951 & 2.71 & 2.11 & 2.27 \\
\rowcolor{Gray}
Tune-A-Video w/ Ours  & \textbf{0.8633} & \textbf{0.9568} & \textbf{3.41} & \textbf{2.88} & \textbf{3.25} \\
\cmidrule{1-6}
ControlVideo           & 0.8686 & 0.9451 & 2.88 & 2.25 & 2.69 \\
\rowcolor{Gray}
ControlVideo w/ Ours    & \textbf{0.8781} & \textbf{0.9590} & \textbf{3.56} & \textbf{3.16} & \textbf{3.40} \\
\bottomrule
\end{tabular}
}
\caption{
Quantitative evaluation of SMA within text-to-image based frameworks.
}
\vspace{-0.4cm}
\label{quantitative-t2i}
\end{table}

\subsection{Quantitative Comparison.}
Quantitative results are detailed in Tab. \ref{quantitative-t2i}, following the metrics introduced in Sec. \ref{sec: quantiative-t2v}.
Across both the Tune-A-Video and ControlVideo frameworks, the integration of SMA improves performance across all five evaluated metrics, notably achieving a substantial advantage in motion accuracy.

\vspace{-0.2cm}
\section{Analysis}
{We explore the impact of SMA by examining motion vectors ($\delta \vb_0$, $\delta \hat{\vb}_0(t)$) and their (amplitude, phase) spectrum in Figure \ref{fig: local ablation} ($t=700$). Figure \ref{fig: local ablation}b indicates that the pixel-space motion vector $\delta \vb_0$ faces frame-wise distortions or inconsistencies. For instance, the motion-independent artifacts, e.g. stair and fence patterns, background texture, etc., persist as distortions in the obtained motion vectors. These are characterized as high-frequency noises in the amplitude spectrum, Fig. \ref{fig: local ablation}a. Motivated by these observations, we prioritize low spatial frequency components during motion distillation to avoid overfitting to motion-independent high-frequency distortions. This representations refinement improves the overall fidelity and removes background distortions (Figure \ref{fig: local ablation}c).}

Moreover, global motion alignment further improves motion transfer. Specifically, without considering the global motion dynamics, existing frameworks occasionally generate ``reversed" motions, i.e. an astronaut skateboarding in an upward direction. This highlights the limitations of conventional frameworks in understanding accurate motion from a single frame of motion vector $\delta \vb_0^n$. In contrast, the proposed global motion alignment effectively mitigates these challenges, ensuring accurate learning of motion patterns. Table \ref{table-ablation} demonstrates the importance of both global/local terms (More qualitative ablation results in appendix).

\begin{table}[htbp]
\centering
\resizebox{\columnwidth}{!}{
\begin{tabular}{@{\extracolsep{0pt}} c c c c c}
\toprule
        & Baseline & Baseline $+\mathcal{L}_\text{local}$  & Baseline $+\mathcal{L}_\text{global}$ & Baseline $+\mathcal{L}_\text{local} +\mathcal{L}_\text{global}$ \\
\midrule
Text-Align & 0.8163 & 0.8382 & 0.8344 & \textbf{0.8427} (\textcolor{teal}{+ 0.026})\\
Frame-Con  & 0.9497 & 0.9613 & 0.9585 & \textbf{0.9659} (\textcolor{teal}{+ 0.016}) \\
\bottomrule
\end{tabular}
}
\caption{{Quantitative ablation of $\mathcal{L}_\text{local}$ and $\mathcal{L}_\text{global}$.}}
\vspace{-0.4cm}
\label{table-ablation}
\end{table}



\section{Conclusion}
We propose Spectral Motion Alignment (SMA), a novel motion distillation framework in spectral domain. We explore the limitations of conventional motion estimation methods: (\textbf{a}) lack of global motion understanding, (\textbf{b}) vulnerability to spatial artifacts. Then, we mitigate these problems by harmonizing both local and global motion alignment and effectively distills motion patterns.

\subsubsection*{Acknowledgments}
This work was supported by Institute of Information $\&$ communications Technology Planning $\&$ Evaluation (IITP) grant funded by the Korea government(MSIT)  
(No.2019-0-00075, Artificial Intelligence Graduate School Program(KAIST), No. RS-2023-00233251, System3 reinforcement learning with high-level brain functions), National Research foundation of Korea(NRF) (**RS-2023-00262527**, RS-2024-00336454, RS-2024-00341805). 

\bibliography{aaai25}

\begin{thebibliography}{43}
\providecommand{\natexlab}[1]{#1}

\bibitem[{Bai et~al.(2024)Bai, He, Wang, Guo, Hu, Liu, and Bian}]{bai2024uniedit}
Bai, J.; He, T.; Wang, Y.; Guo, J.; Hu, H.; Liu, Z.; and Bian, J. 2024.
\newblock UniEdit: A Unified Tuning-Free Framework for Video Motion and Appearance Editing.
\newblock \emph{arXiv preprint arXiv:2402.13185}.

\bibitem[{Bain et~al.(2021)Bain, Nagrani, Varol, and Zisserman}]{Bain21}
Bain, M.; Nagrani, A.; Varol, G.; and Zisserman, A. 2021.
\newblock Frozen in Time: A Joint Video and Image Encoder for End-to-End Retrieval.
\newblock In \emph{IEEE International Conference on Computer Vision}.

\bibitem[{Ceylan, Huang, and Mitra(2023)}]{ceylan2023pix2video}
Ceylan, D.; Huang, C.-H.~P.; and Mitra, N.~J. 2023.
\newblock Pix2video: Video editing using image diffusion.
\newblock In \emph{Proceedings of the IEEE/CVF International Conference on Computer Vision}, 23206--23217.

\bibitem[{Chen et~al.(2023{\natexlab{a}})Chen, Xia, He, Zhang, Cun, Yang, Xing, Liu, Chen, Wang et~al.}]{chen2023videocrafter1}
Chen, H.; Xia, M.; He, Y.; Zhang, Y.; Cun, X.; Yang, S.; Xing, J.; Liu, Y.; Chen, Q.; Wang, X.; et~al. 2023{\natexlab{a}}.
\newblock Videocrafter1: Open diffusion models for high-quality video generation.
\newblock \emph{arXiv preprint arXiv:2310.19512}.

\bibitem[{Chen et~al.(2023{\natexlab{b}})Chen, Wu, Xie, Wu, Li, Xia, Xiao, and Lin}]{chen2023control}
Chen, W.; Wu, J.; Xie, P.; Wu, H.; Li, J.; Xia, X.; Xiao, X.; and Lin, L. 2023{\natexlab{b}}.
\newblock Control-A-Video: Controllable Text-to-Video Generation with Diffusion Models.
\newblock \emph{arXiv preprint arXiv:2305.13840}.

\bibitem[{Efron(2011)}]{efron2011tweedie}
Efron, B. 2011.
\newblock Tweedie’s formula and selection bias.
\newblock \emph{Journal of the American Statistical Association}, 106(496): 1602--1614.

\bibitem[{Hessel et~al.(2021)Hessel, Holtzman, Forbes, Bras, and Choi}]{hessel2021clipscore}
Hessel, J.; Holtzman, A.; Forbes, M.; Bras, R.~L.; and Choi, Y. 2021.
\newblock Clipscore: A reference-free evaluation metric for image captioning.
\newblock \emph{arXiv preprint arXiv:2104.08718}.

\bibitem[{Ho et~al.(2022{\natexlab{a}})Ho, Chan, Saharia, Whang, Gao, Gritsenko, Kingma, Poole, Norouzi, Fleet et~al.}]{ho2022imagen}
Ho, J.; Chan, W.; Saharia, C.; Whang, J.; Gao, R.; Gritsenko, A.; Kingma, D.~P.; Poole, B.; Norouzi, M.; Fleet, D.~J.; et~al. 2022{\natexlab{a}}.
\newblock Imagen video: High definition video generation with diffusion models.
\newblock \emph{arXiv preprint arXiv:2210.02303}.

\bibitem[{Ho, Jain, and Abbeel(2020)}]{ho2020denoising}
Ho, J.; Jain, A.; and Abbeel, P. 2020.
\newblock Denoising diffusion probabilistic models.
\newblock \emph{Advances in neural information processing systems}, 33: 6840--6851.

\bibitem[{Ho et~al.(2022{\natexlab{b}})Ho, Salimans, Gritsenko, Chan, Norouzi, and Fleet}]{ho2022video}
Ho, J.; Salimans, T.; Gritsenko, A.; Chan, W.; Norouzi, M.; and Fleet, D.~J. 2022{\natexlab{b}}.
\newblock Video diffusion models.
\newblock \emph{arXiv:2204.03458}.

\bibitem[{Hu et~al.(2021)Hu, Shen, Wallis, Allen-Zhu, Li, Wang, Wang, and Chen}]{hu2021lora}
Hu, E.~J.; Shen, Y.; Wallis, P.; Allen-Zhu, Z.; Li, Y.; Wang, S.; Wang, L.; and Chen, W. 2021.
\newblock Lora: Low-rank adaptation of large language models.
\newblock \emph{arXiv preprint arXiv:2106.09685}.

\bibitem[{Hu and Xu(2023)}]{hu2023videocontrolnet}
Hu, Z.; and Xu, D. 2023.
\newblock Videocontrolnet: A motion-guided video-to-video translation framework by using diffusion model with controlnet.
\newblock \emph{arXiv preprint arXiv:2307.14073}.

\bibitem[{Jeong et~al.(2024)Jeong, Chang, Park, and Ye}]{jeong2024dreammotion}
Jeong, H.; Chang, J.; Park, G.~Y.; and Ye, J.~C. 2024.
\newblock DreamMotion: Space-Time Self-Similarity Score Distillation for Zero-Shot Video Editing.
\newblock \emph{arXiv preprint arXiv:2403.12002}.

\bibitem[{Jeong, Park, and Ye(2023)}]{jeong2023vmc}
Jeong, H.; Park, G.~Y.; and Ye, J.~C. 2023.
\newblock VMC: Video Motion Customization using Temporal Attention Adaption for Text-to-Video Diffusion Models.
\newblock \emph{arXiv preprint arXiv:2312.00845}.

\bibitem[{Jeong and Ye(2023)}]{jeong2023ground}
Jeong, H.; and Ye, J.~C. 2023.
\newblock Ground-a-video: Zero-shot grounded video editing using text-to-image diffusion models.
\newblock \emph{arXiv preprint arXiv:2310.01107}.

\bibitem[{Khachatryan et~al.(2023)Khachatryan, Movsisyan, Tadevosyan, Henschel, Wang, Navasardyan, and Shi}]{khachatryan2023text2video}
Khachatryan, L.; Movsisyan, A.; Tadevosyan, V.; Henschel, R.; Wang, Z.; Navasardyan, S.; and Shi, H. 2023.
\newblock Text2video-zero: Text-to-image diffusion models are zero-shot video generators.
\newblock In \emph{Proceedings of the IEEE/CVF International Conference on Computer Vision}, 15954--15964.

\bibitem[{Kim et~al.(2024)Kim, Lee, Park, Kim, Lee, Kim, and Yoo}]{kim2024hybrid}
Kim, K.; Lee, H.; Park, J.; Kim, S.; Lee, K.; Kim, S.; and Yoo, J. 2024.
\newblock Hybrid Video Diffusion Models with 2D Triplane and 3D Wavelet Representation.
\newblock \emph{arXiv preprint arXiv:2402.13729}.

\bibitem[{Li et~al.(2023)Li, Liu, Wu, Mu, Yang, Gao, Li, and Lee}]{li2023gligen}
Li, Y.; Liu, H.; Wu, Q.; Mu, F.; Yang, J.; Gao, J.; Li, C.; and Lee, Y.~J. 2023.
\newblock Gligen: Open-set grounded text-to-image generation.
\newblock In \emph{Proceedings of the IEEE/CVF Conference on Computer Vision and Pattern Recognition}, 22511--22521.

\bibitem[{Magarey and Kingsbury(1998)}]{magarey1998motion}
Magarey, J.; and Kingsbury, N. 1998.
\newblock Motion estimation using a complex-valued wavelet transform.
\newblock \emph{IEEE Transactions on Signal Processing}, 46(4): 1069--1084.

\bibitem[{Pont-Tuset et~al.(2017)Pont-Tuset, Perazzi, Caelles, Arbel{\'a}ez, Sorkine-Hornung, and Van~Gool}]{pont20172017}
Pont-Tuset, J.; Perazzi, F.; Caelles, S.; Arbel{\'a}ez, P.; Sorkine-Hornung, A.; and Van~Gool, L. 2017.
\newblock The 2017 davis challenge on video object segmentation.
\newblock \emph{arXiv preprint arXiv:1704.00675}.

\bibitem[{Qi et~al.(2023)Qi, Cun, Zhang, Lei, Wang, Shan, and Chen}]{qi2023fatezero}
Qi, C.; Cun, X.; Zhang, Y.; Lei, C.; Wang, X.; Shan, Y.; and Chen, Q. 2023.
\newblock Fatezero: Fusing attentions for zero-shot text-based video editing.
\newblock \emph{arXiv preprint arXiv:2303.09535}.

\bibitem[{Ren et~al.(2024)Ren, Zhou, Yang, Shi, Liu, Liu, Kwon, and Shrivastava}]{ren2024customize}
Ren, Y.; Zhou, Y.; Yang, J.; Shi, J.; Liu, D.; Liu, F.; Kwon, M.; and Shrivastava, A. 2024.
\newblock Customize-A-Video: One-Shot Motion Customization of Text-to-Video Diffusion Models.
\newblock \emph{arXiv preprint arXiv:2402.14780}.

\bibitem[{Rombach et~al.(2022)Rombach, Blattmann, Lorenz, Esser, and Ommer}]{rombach2022high}
Rombach, R.; Blattmann, A.; Lorenz, D.; Esser, P.; and Ommer, B. 2022.
\newblock High-resolution image synthesis with latent diffusion models.
\newblock In \emph{Proceedings of the IEEE/CVF conference on computer vision and pattern recognition}, 10684--10695.

\bibitem[{Secker and Taubman(2002)}]{secker2002highly}
Secker, A.; and Taubman, D. 2002.
\newblock Highly scalable video compression using a lifting-based 3D wavelet transform with deformable mesh motion compensation.
\newblock In \emph{Proceedings. International Conference on Image Processing}, volume~3, 749--752. IEEE.

\bibitem[{Sohl-Dickstein et~al.(2015)Sohl-Dickstein, Weiss, Maheswaranathan, and Ganguli}]{sohl2015deep}
Sohl-Dickstein, J.; Weiss, E.; Maheswaranathan, N.; and Ganguli, S. 2015.
\newblock Deep unsupervised learning using nonequilibrium thermodynamics.
\newblock In \emph{International conference on machine learning}, 2256--2265. PMLR.

\bibitem[{Song, Meng, and Ermon(2020)}]{song2020denoising}
Song, J.; Meng, C.; and Ermon, S. 2020.
\newblock Denoising diffusion implicit models.
\newblock \emph{arXiv preprint arXiv:2010.02502}.

\bibitem[{Sterling(2023)}]{zeroscope}
Sterling, S. 2023.
\newblock Zeroscope.
\newblock \url{https://huggingface.co/cerspense/zeroscope_v2_576w}.

\bibitem[{Tang et~al.(2023)Tang, Jia, Wang, Phoo, and Hariharan}]{tang2023emergent}
Tang, L.; Jia, M.; Wang, Q.; Phoo, C.~P.; and Hariharan, B. 2023.
\newblock Emergent correspondence from image diffusion.
\newblock \emph{Advances in Neural Information Processing Systems}, 36: 1363--1389.

\bibitem[{Wang et~al.(2023{\natexlab{a}})Wang, Yuan, Chen, Zhang, Wang, and Zhang}]{wang2023modelscope}
Wang, J.; Yuan, H.; Chen, D.; Zhang, Y.; Wang, X.; and Zhang, S. 2023{\natexlab{a}}.
\newblock Modelscope text-to-video technical report.
\newblock \emph{arXiv preprint arXiv:2308.06571}.

\bibitem[{Wang et~al.(2023{\natexlab{b}})Wang, Chen, Ma, Zhou, Huang, Wang, Yang, He, Yu, Yang et~al.}]{wang2023lavie}
Wang, Y.; Chen, X.; Ma, X.; Zhou, S.; Huang, Z.; Wang, Y.; Yang, C.; He, Y.; Yu, J.; Yang, P.; et~al. 2023{\natexlab{b}}.
\newblock Lavie: High-quality video generation with cascaded latent diffusion models.
\newblock \emph{arXiv preprint arXiv:2309.15103}.

\bibitem[{Wei et~al.(2023)Wei, Zhang, Qing, Yuan, Liu, Liu, Zhang, Zhou, and Shan}]{wei2023dreamvideo}
Wei, Y.; Zhang, S.; Qing, Z.; Yuan, H.; Liu, Z.; Liu, Y.; Zhang, Y.; Zhou, J.; and Shan, H. 2023.
\newblock Dreamvideo: Composing your dream videos with customized subject and motion.
\newblock \emph{arXiv preprint arXiv:2312.04433}.

\bibitem[{Williams et~al.(2024)Williams, Falck, Deligiannidis, Holmes, Doucet, and Syed}]{williams2024unified}
Williams, C.; Falck, F.; Deligiannidis, G.; Holmes, C.~C.; Doucet, A.; and Syed, S. 2024.
\newblock A Unified Framework for U-Net Design and Analysis.
\newblock \emph{Advances in Neural Information Processing Systems}, 36.

\bibitem[{Wu et~al.(2023)Wu, Ge, Wang, Lei, Gu, Shi, Hsu, Shan, Qie, and Shou}]{wu2023tune}
Wu, J.~Z.; Ge, Y.; Wang, X.; Lei, S.~W.; Gu, Y.; Shi, Y.; Hsu, W.; Shan, Y.; Qie, X.; and Shou, M.~Z. 2023.
\newblock Tune-a-video: One-shot tuning of image diffusion models for text-to-video generation.
\newblock In \emph{Proceedings of the IEEE/CVF International Conference on Computer Vision}, 7623--7633.

\bibitem[{Yang et~al.(2022)Yang, Liu, Liu, Gu, Cao, and Li}]{yang2022delving}
Yang, G.; Liu, W.; Liu, X.; Gu, X.; Cao, J.; and Li, J. 2022.
\newblock Delving into the frequency: Temporally consistent human motion transfer in the fourier space.
\newblock In \emph{Proceedings of the 30th ACM International Conference on Multimedia}, 1156--1166.

\bibitem[{Yang et~al.(2024)Yang, Hou, Huang, Ma, Wan, Zhang, Chen, and Liao}]{yang2024direct}
Yang, S.; Hou, L.; Huang, H.; Ma, C.; Wan, P.; Zhang, D.; Chen, X.; and Liao, J. 2024.
\newblock Direct-a-Video: Customized Video Generation with User-Directed Camera Movement and Object Motion.
\newblock \emph{arXiv preprint arXiv:2402.03162}.

\bibitem[{Yatim et~al.(2023)Yatim, Fridman, Tal, Kasten, and Dekel}]{yatim2023space}
Yatim, D.; Fridman, R.; Tal, O.~B.; Kasten, Y.; and Dekel, T. 2023.
\newblock Space-Time Diffusion Features for Zero-Shot Text-Driven Motion Transfer.
\newblock \emph{arXiv preprint arXiv:2311.17009}.

\bibitem[{Ye, Han, and Cha(2018)}]{ye2018deep}
Ye, J.~C.; Han, Y.; and Cha, E. 2018.
\newblock Deep convolutional framelets: A general deep learning framework for inverse problems.
\newblock \emph{SIAM Journal on Imaging Sciences}, 11(2): 991--1048.

\bibitem[{Yoo et~al.(2019)Yoo, Uh, Chun, Kang, and Ha}]{yoo2019photorealistic}
Yoo, J.; Uh, Y.; Chun, S.; Kang, B.; and Ha, J.-W. 2019.
\newblock Photorealistic style transfer via wavelet transforms.
\newblock In \emph{Proceedings of the IEEE/CVF international conference on computer vision}, 9036--9045.

\bibitem[{Zhang et~al.(2023{\natexlab{a}})Zhang, Wu, Liu, Zhao, Ran, Gu, Gao, and Shou}]{zhang2023show}
Zhang, D.~J.; Wu, J.~Z.; Liu, J.-W.; Zhao, R.; Ran, L.; Gu, Y.; Gao, D.; and Shou, M.~Z. 2023{\natexlab{a}}.
\newblock Show-1: Marrying pixel and latent diffusion models for text-to-video generation.
\newblock \emph{arXiv preprint arXiv:2309.15818}.

\bibitem[{Zhang, Rao, and Agrawala(2023)}]{zhang2023adding}
Zhang, L.; Rao, A.; and Agrawala, M. 2023.
\newblock Adding conditional control to text-to-image diffusion models.
\newblock In \emph{Proceedings of the IEEE/CVF International Conference on Computer Vision}, 3836--3847.

\bibitem[{Zhang et~al.(2023{\natexlab{b}})Zhang, Wei, Jiang, Zhang, Zuo, and Tian}]{zhang2023controlvideo}
Zhang, Y.; Wei, Y.; Jiang, D.; Zhang, X.; Zuo, W.; and Tian, Q. 2023{\natexlab{b}}.
\newblock Controlvideo: Training-free controllable text-to-video generation.
\newblock \emph{arXiv preprint arXiv:2305.13077}.

\bibitem[{Zhao et~al.(2023{\natexlab{a}})Zhao, Wang, Bao, Li, and Zhu}]{zhao2023controlvideo}
Zhao, M.; Wang, R.; Bao, F.; Li, C.; and Zhu, J. 2023{\natexlab{a}}.
\newblock ControlVideo: Adding Conditional Control for One Shot Text-to-Video Editing.
\newblock \emph{arXiv preprint arXiv:2305.17098}.

\bibitem[{Zhao et~al.(2023{\natexlab{b}})Zhao, Gu, Wu, Zhang, Liu, Wu, Keppo, and Shou}]{zhao2023motiondirector}
Zhao, R.; Gu, Y.; Wu, J.~Z.; Zhang, D.~J.; Liu, J.; Wu, W.; Keppo, J.; and Shou, M.~Z. 2023{\natexlab{b}}.
\newblock Motiondirector: Motion customization of text-to-video diffusion models.
\newblock \emph{arXiv preprint arXiv:2310.08465}.

\end{thebibliography}

\appendix

\onecolumn

\section{Pseudo Training Algorithm}
In our work, we adopt the notation and expressions mostly from \cite{jeong2023vmc} for the preliminaries section and pseudo-code, due to its relevance to our focus on denoised motion vector estimates. We interchangeably use $\hat{\vb}_0^{1:N}(t)$ and $\hat{\vb}_0(t)$ in the main paper and Algorithm 1. While Algorithm 1 generalizes parameter $\theta$, each video diffusion model incorporates specific parameters such as $\theta_{\text{TA}}$ \cite{jeong2023vmc} and $\theta_{\text{LoRA}}$ \cite{zhao2023motiondirector}. 

\begin{algorithm}
\label{pseudocode}
	\caption{\textbf{Spectral Motion Alignment (SMA)}} 
	\begin{algorithmic}[1]
	    \State {\bfseries Input:} $N$-frame input video sequence $(\vb_0^n)_{n \in \{1, \dots, N\}}$, training prompt $\Pc$, textual encoder $\psi$, Training iterations $M$, Video diffusion models parameterized by $\theta$. 
     
        \State {\bfseries Output:} Fine-tuned video diffusion models \(\theta^*\). \\

            \For {$step=1$ \textbf{to} $M$}
                \State Sample timestep $t \in [0, T]$ and Gaussian noise $\epsilonb_t^{1:N}$, where \(\epsilonb_t^n \in \mathbb{R}^d \sim \Nc(0, I)\)
                \State Prepare text embeddings $c = \psi(\Pc)$ \\

                \State \textit{1. Denoised motion vector estimation}
                \State $\vb_t^{1:N} = \sqrt{\bar{\alpha}_t} \vb_0^{1:N} + \sqrt{1 - \bar{\alpha}_t} \epsilonb_t^{1:N}$. 
                \State $\hat{\vb}_0^{1:N}(t) = \frac{1}{\sqrt{\alphabar_t}} \big( \vb_t^{1:N} - \sqrt{1-\alphabar_t} \epsilonb_{\theta} (\vb_t^{1:N}, t, c) \big)$. \\

                \State \textit{2. Global motion alignment}
                \State Conduct 1D DWT for each $s$-th pixel in $\delta \vb_0, \delta \hat{\vb}_0(t)$ with Haar wavelet.
                \State $\ell_{\text{global}}(\delta \vb_0, \delta \hat{\vb}_0(t)) = \mathbb{E}_{t, s, j, k} \Big[ \| \Wc_{\delta \vb_{0, s}}(j, k) - \Wc_{\delta \hat{\vb}_{0, s}(t)}(j, k) \|_1 \Big]$. \\

                \State \textit{3. Local motion refinement}
                \State Obtain amplitude and phase spectrum for $\delta \vb_0^n$ as $| \mathcal{F}_{\delta \vb_0^n} (a, b) |, \angle \mathcal{F}_{\delta \vb_0^n} (a, b)$.
                \State $\ell_{\local}^{a}(\delta \vb_0^n, \delta \hat{\vb}_0^n(t)) = \mathbb{E}_{t, n, a, b} \Big[ \Wc(a, b) * \| |\mathcal{F}_{\delta \vb_0^n} (a, b)| - |\mathcal{F}_{\delta \hat{\vb}_0^n(t)} (a, b)| \|_1  \Big]$.
                \State $\ell_{\local}^{p}(\delta \vb_0^n, \delta \hat{\vb}_0^n(t)) = \mathbb{E}_{t, n, a, b} \Big[ \Wc(a, b) * \| \angle \mathcal{F}_{\delta \vb_0^n} (a, b) - \angle \mathcal{F}_{\delta \hat{\vb}_0^n(t)} (a, b) \|_1  \Big]$. \\

                \State \textit{4. Overall optimization}
                \State $\theta^* = \arg \min_\theta \mathbb E_{t, n, \epsilonb_t^n, \epsilonb_t^{n+1}} \Big[\ell_{SMA}(\delta \vb_0, \delta \hat{\vb}_0(t)\big)\Big]$ ($\ell_{SMA}$: objective in eq(20)).
            \EndFor
            
	\end{algorithmic} 
\end{algorithm}

\section{Related Work}
\subsection{Diffusion-based Video Editing}
There has been considerable progress in adapting the achievements of diffusion-based image editing for video generative tasks.
Compared to text-conditioned image generation, creating videos based solely on text introduces the complex challenge of producing temporally consistent and natural motion.
In the absence of publicly available text-to-video diffusion models, Tune-A-Video \cite{wu2023tune} was at the forefront of one-shot based video editing.
It proposes to inflate image diffusion model to pseudo video diffusion model by appending temporal modules to image diffusion model \cite{rombach2022high} and reprogramming spatial self-attention to spatio-temporal self-attention.
Following this adaptation, the attention modules' query projection matrices are fine-tuned on the input video.
To eliminate the need for customizing model weights for every new video, various zero-shot editing methods have been developed.
One approach involves guiding the generation process with attention maps, such as the inekction of self-attention maps obtained from the input video \cite{ceylan2023pix2video, qi2023fatezero}. 
Another prevalent method integrates explicit structural cues, like depth or edge maps, into the reverse diffusion process.
For instance, ControlNet \cite{zhang2023adding} has been adapted for the video domain, facilitating the creation of structurally consistent frames in video generation \cite{khachatryan2023text2video} and translation tasks \cite{hu2023videocontrolnet, zhang2023controlvideo}.
Furthermore, GLIGEN's \cite{li2023gligen} adaptation to the video domain by Ground-A-Video \cite{jeong2023ground} demonstrates the use of both spatially-continuous depth map and spatially-discrete bounding box conditions, achieving multi-attribute editing of videos in a zero-shot manner.

The advent of open-source text-to-video diffusion models \cite{wang2023modelscope, zeroscope, chen2023videocrafter1, wang2023lavie} has spurred research into separating, altering, and combining the appearance and motion elements of videos \cite{zhao2023motiondirector, jeong2023vmc, wei2023dreamvideo, bai2024uniedit, yang2024direct, yatim2023space, ren2024customize, jeong2024dreammotion}.
MotionDirector \cite{zhao2023motiondirector} and DreamVideo \cite{wei2023dreamvideo} have each suggested approaches for dividing fine-tuning processes into distinct learning phases for subject appearance and temporal motion, utilizing efficient fine-tuning methods.
On the other hand, VMC \cite{jeong2023vmc} focuses on distilling the motion within a video by calculating the residual vectors between consecutive frames.
In their work, they fine-tune temporal attention layers within cascaded video diffusion models to synchronize the ground-truth motion vector with the denoised motion vector estimate, successfully generating videos that replicate the motion pattern of an input video within diverse visual scenarios.
Similarly, \cite{yatim2023space} introduces a space-time feature loss that constructs self-similarity matrices based on the differences in attention features between frames. This approach aims to minimize the discrepancy in self-similarity between the input and output videos.

\subsection{Frequency-aware Visual Generation}
Spectral analysis plays a pivotal role in the domain of visual understanding and generation, offering insights into the temporal-spatial structure of pixel-domain frames through frequency-domain signals. \cite{magarey1998motion} introduced a hierarchical motion estimation algorithm employing complex discrete wavelet transforms, effectively utilizing phase differences among subbands to indicate local translations within video frames. \cite{secker2002highly} enhanced scalable video compression through motion-compensated wavelet transforms, integrating a continuous deformable mesh motion model to achieve superior compression efficiency and motion representation.

Furthermore, these spectral insights have been instrumental in refining algorithms and deepening architectural understanding, particularly within the contexts of U-Net and autoencoder frameworks. \cite{ye2018deep} provided a groundbreaking reinterpretation of deep learning architectures for image reconstruction, establishing a connection between deep learning and classical signal processing theories, including wavelets and compressed sensing. \cite{yoo2019photorealistic} developed a wavelet-based correction method, WCT2, to augment photorealism in style transfer, leveraging whitening and coloring transforms to preserve structural integrity and statistics within the VGG feature space.

In the realm of U-Net in diffusion models, \cite{williams2024unified} highlighted the rapid dominance of noise over high-frequency information in residual U-Net denoisers. Concurrently, \cite{kim2024hybrid} proposed the Hybrid Video Diffusion Model (HVDM), a novel architecture that captures spatiotemporal dependencies using a disentangled representation combining 2D projection and 3D convolutions with wavelet decomposition. While \cite{kim2024hybrid} aims to \textit{reconstruct} input video with frequency matching loss, our approach does not aim to reconstruct input, focusing instead on learning motion dynamics through motion vectors, thereby distinguishing our method within the landscape of spectral analysis applications in motion estimation and transfer.

\section{Experimental details}
For spectral global motion alignment, we mainly use Haar wavelets with the number of levels $l=3$ for 8-frame videos and $l=4$ for 16-frame videos. We use \texttt{DWT1DForward} function from the PyTorch package\footnote{\url{https://pytorch-wavelets.readthedocs.io/en/latest/index.html}}. For spectral local motion refinement, we fix $\delta=0.05$ in frequency domain weighting $\omega(a,b)$. We set $\lambda_g=0.4$ and $\lambda_l=0.2$ for many cases, where we recommend to fine-tune $\lambda_g$ in a range of $[0.2, 0.5]$, and $\lambda_g$ in a range of $[0.1, 0.3]$. We follow other configurations, e.g. optimization algorithm, learning rate, training steps, etc., from the original motion transfer frameworks for a fair comparisons. Code is included as a supplementary material.

\section{Additional analysis}
\subsubsection{Ablation study.} 
\begin{figure}[htb]
    \centering
    \includegraphics[width=\textwidth]{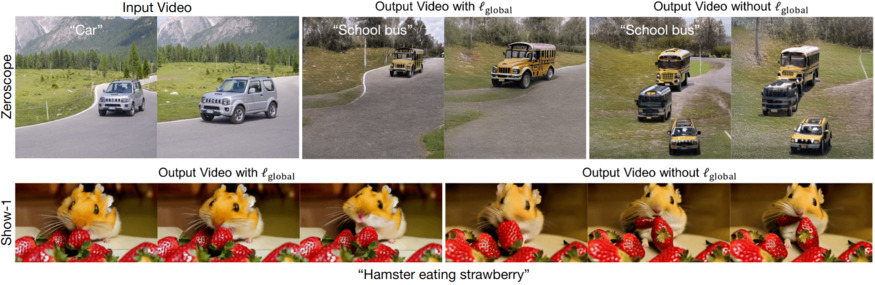}
    \caption{Ablation study on global alignment. Global motion alignment facilitates motion transfer.}
    \label{supple_fig: ablation}
\end{figure}

We first provide additional qualitative ablation results of global alignment in Fig. \ref{supple_fig: ablation} (baseline: VMC). Combined with quantiative ablation studies in the main paper, these demonstrates the effectiveness of both global and local motion alignment. Please refer to the Figure 6 in the main paper for more analysis on the local term.

\subsubsection{Training progress.} To analyze the training progress further, we visualize the intermediate customization results with and without proposed spectral motion alignment. Notably, Fig. \ref{supple_fig: analysis}a illustrates the gradual improvements of spatial layout attributed to the proposed global/local motion alignment. Specifically, our layout progressively aligns better as training proceeds. Both global and local terms contribute to this alignment as both match motion vectors $\delta \boldsymbol{v}_0$ encoding structural movement information.

\subsubsection{Alternative frequency-domain approaches.} 
\begin{figure}[htb]
    \centering
    \includegraphics[width=\textwidth]{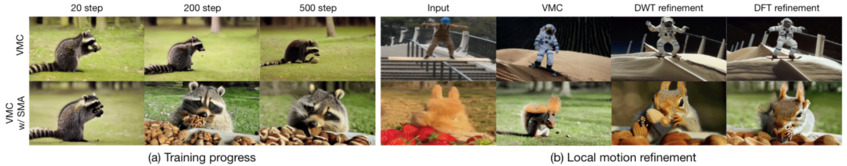}
    \caption{(\textbf{a}) Visualization of the training progress with and without local term (Global motion alignment leads to similar trends). (\textbf{b}) Comparisons of local motion refinement with 2D DWT and 2D DFT (originally proposed).}
    \label{supple_fig: analysis}
\end{figure}

Our framework highlights two key insights: (a) susceptibility of 2D motion information in $\delta \boldsymbol{v}_0$ to spatial/motion artifacts, and (b) potential benefits of multi-scale decomposition for refining motion representation. In this context, both multi-scale Fourier and wavelet analyses offer theoretically viable options for spectral motion refinement. Demonstrating this, we use 2D Discrete Wavelet Transform (DWT) with Daubechies 3 wavelet, to obtain wavelet coefficients of ground truth and predicted motion vectors. Then each motion vector is reconstructed from these coefficients, excluding the finest high-frequency detail coefficients. Figure \ref{supple_fig: analysis}b illustrates that DWT-based refinement also significantly improves motion transfer compared to baseline (VMC), particularly by prioritizing low-frequency components of local motion encoding core motion information. While both methods show notable improvements, DFT outperforms DWT, with DWT requiring more fine-tuning of hyperparameters (e.g., the number of levels depending on spatial resolution, wavelet type, etc). Thus, we opt for DFT in refinement.

\section{Additional Results}
\begin{figure}[htb]
    \centering
    \includegraphics[width=\textwidth]{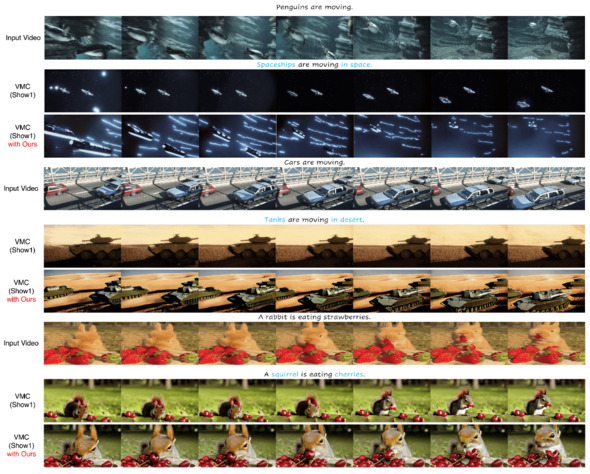}
    \caption{Additional comparison within VMC framework, deployed on Show-1 Cascade.}
    \label{supple_fig: vmc_show1}
\end{figure}

\begin{figure}[htb]
    \centering 
    \includegraphics[width=\textwidth]{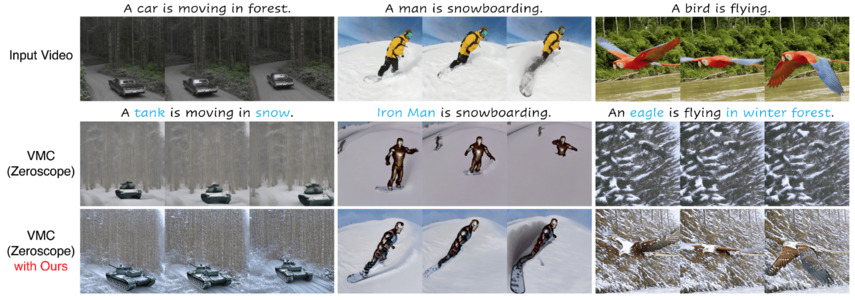}
    \caption{Additional comparison within VMC method, implemented on Zeroscope T2V.}
    \label{supple_fig: vmc_zeroscope}
\end{figure}

\begin{figure}[htb]
    \centering 
    \includegraphics[width=\textwidth]{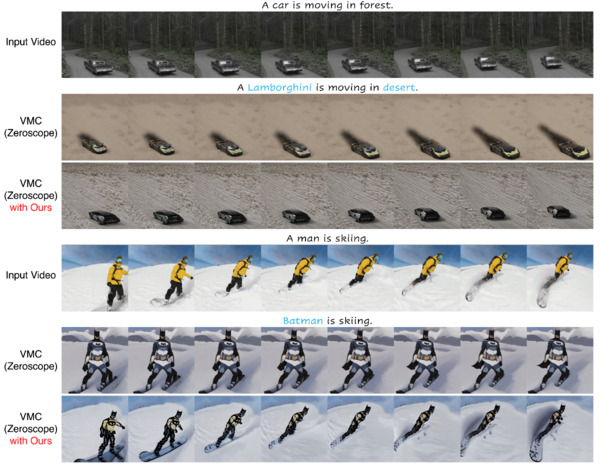}
    \caption{Additional comparison within VMC method, implemented on Zeroscope T2V.}
    \label{supple_fig: vmc_zeroscope2}
\end{figure}

\begin{figure}[htb]
    \centering 
    \includegraphics[width=\textwidth]{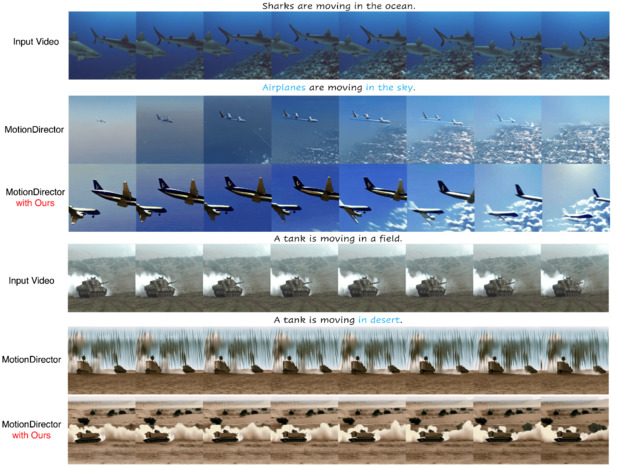}
    \caption{Additional comparison within MotionDirector method.}
    \label{supple_fig: motiondirector}
\end{figure}

\begin{figure}[htb]
    \centering
    \includegraphics[width=\textwidth]{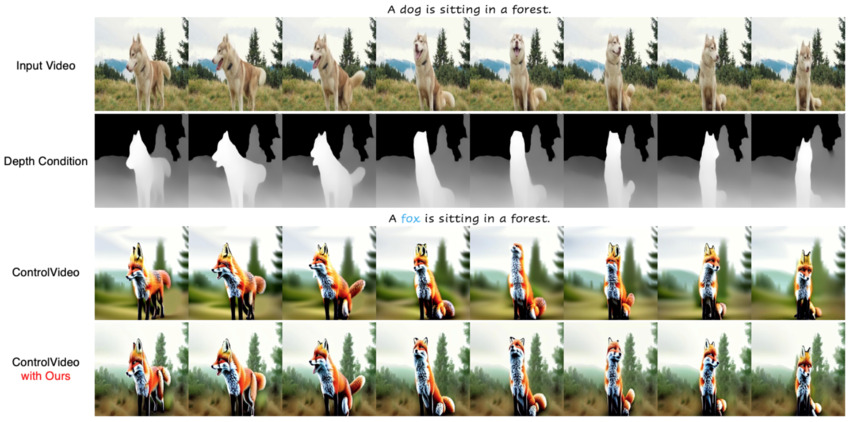}
    \caption{Additional comparison within ControlVideo frameworks.}
    \label{supple_fig: cav}
\end{figure}

This section provides additional qualitative comparisons of SMA across different baseline approaches. Figures \ref{supple_fig: vmc_show1} and \ref{supple_fig: vmc_zeroscope}, \ref{supple_fig: vmc_zeroscope2} delve into VMC's \cite{jeong2023vmc} capabilities in customizing motion, showcasing outcomes with SMA and without its application. In Figure \ref{supple_fig: motiondirector}, we compare the efficacy of MotionDirector \cite{zhao2023motiondirector} in transferring motion, both with SMA integrated and without. 
Additionally, Figure \ref{supple_fig: cav} contrasts the ability of ControlVideo \cite{wu2023tune} to replicate the original motion, examining the impact of incorporating SMA.

\end{document}